\def\year{2020}\relax
\documentclass[letterpaper]{article} % DO NOT CHANGE THIS
\usepackage{aaai20}  % DO NOT CHANGE THIS
\usepackage{times}  % DO NOT CHANGE THIS
\usepackage{helvet} % DO NOT CHANGE THIS
\usepackage{courier}  % DO NOT CHANGE THIS
\usepackage[hyphens]{url}  % DO NOT CHANGE THIS
\usepackage{graphicx} % DO NOT CHANGE THIS
\usepackage{amsfonts}
\usepackage{multirow}
\usepackage{amssymb}
\usepackage{amsmath}
\usepackage{graphicx}  % DO NOT CHANGE THIS
\title{Multi-Task Driven Feature Models for Thermal Infrared Tracking }

\author{Qiao Liu,\textsuperscript{\rm 1}  Xin Li,\textsuperscript{\rm 1}\thanks{Qiao Liu and Xin Li contribute equally.} Zhenyu He,\textsuperscript{\rm 1, \rm 3}\thanks{Zhenyu He is the corresponding author.} Nana Fan,\textsuperscript{\rm 1} Di Yuan,\textsuperscript{\rm 1} Wei Liu,\textsuperscript{\rm 2, \rm3} Yongsheng Liang\textsuperscript{\rm 1}\\ % All authors must be in the same font size and format. Use \Large and \textbf to achieve this result when breaking a line
\textsuperscript{\rm 1}Harbin Institute of Technology, Shenzhen \\
\textsuperscript{\rm 2}Shenzhen Institute of Information Technology\\
\textsuperscript{\rm 3}Peng Cheng Laboratory\\
liuqiao@stu.hit.edu.cn, \{xinlihitsz, nanafanhit, dyuanhit\}@gmail.com\\ \{zhenyuhe, liangyongsheng\}@hit.edu.cn, liuwei@sziit.edu.cn % email address must be in roman text type, not monospace or sans serif
}
\begin{document}

\maketitle

\begin{abstract}
Existing deep Thermal InfraRed (TIR) trackers usually use the feature models of RGB trackers for representation.
However, these feature models learned on RGB images are neither effective in representing TIR objects nor taking fine-grained TIR information into consideration.
To this end, we develop a multi-task framework to learn the TIR-specific discriminative features and fine-grained correlation features for TIR tracking.
Specifically,
we first use an auxiliary classification network to guide the generation of TIR-specific discriminative features for distinguishing the TIR objects belonging to different classes.
Second, we design a fine-grained aware module to capture more subtle information for distinguishing the TIR objects belonging to the same class.
These two kinds of features complement each other and recognize TIR objects in the levels of inter-class and intra-class respectively.
These two feature models are learned using a multi-task matching framework and are jointly optimized on the TIR tracking task.
In addition, we develop a large-scale TIR training dataset to train the network for adapting the model to the TIR domain.
Extensive experimental results on three benchmarks show that the proposed algorithm achieves a relative gain of $10\%$ over the baseline and performs favorably against the state-of-the-art methods. Codes and the proposed TIR dataset are available at~\url{https://github.com/QiaoLiuHit/MMNet}.
\end{abstract}

\section{Introduction}
%background
TIR object tracking is an important task in artificial intelligence. It has been widely used in maritime rescue, video surveillance, and driver assistance at night~\cite{TC} as it can track the object in total darkness.
Despite much progress, TIR tracking still faces several challenging problems, such as distractor, occlusion, size change, and thermal cross~\cite{PTB-TIR}.

%review
Inspired by the success of Convolution Neural Networks (CNNs) in visual tracking, there are several attempts to use CNNs to improve the performance of TIR trackers.
These methods can be roughly divided into two categories, deep feature based TIR trackers and matching-based deep TIR trackers.
Deep feature based TIR trackers, e.g., DSST-tir~\cite{16KTIR}, MCFTS~\cite{MCFTS}, and LMSCO~\cite{LMSCO}, use a pre-trained classification network for extracting deep features and then integrate them into conventional trackers.
Despite the demonstrated success, their performance is limited by the pre-trained deep features which are learned from RGB images and are less effective in representing TIR objects.
Matching-based deep TIR tracking methods, e.g., HSSNet~\cite{HSSNet} and MLSSNet~\cite{MLSSNet}, cast tracking as a matching problem and train a matching network off-line for online tracking.
These methods receive much attention recently because of their high efficiency and simplicity.
%problem
However, they are also limited by the weak discriminative capacity of the learned features due to the following reasons.
First, they do not learn how to separate samples belonging to different classes, namely, the learned features are sensitive to all semantic objects.
Second, their features are insensitive to similar objects as they are usually learned on a global semantic feature space without fine-grained information.
Noting that fine-grained information is crucial for distinguishing TIR objects as similar semantic patterns are generated from intra-class TIR objects.
Third, their features are often learned from RGB or small TIR datasets, which do not learn the specific patterns of TIR objects.

%solution
To address the above-mentioned issues, we propose to learn TIR-specific discriminative features and fine-grained correlation features.
Specifically, we use a classification network, targeting at distinguishing TIR objects from different classes, to guide the generation of the TIR-specific discriminative feature.
In addition, we design a fine-grained aware network, which consists of a holistic correlation and pixel-level correlation modules, for obtaining the fine-grained correlation features.
When the TIR-specific discriminative features are not able to distinguish similar distractors, the fine-grained correlation feature provides more detailed information for distinguishing them.

To integrate these two complemental features effectively, we design a multi-task matching framework for learning them simultaneously.
To adapt the feature model to the TIR domain better, we construct a large-scale TIR image sequence dataset to train the proposed network.
The dataset includes $30$ classes, over $1,100$ image sequences, over $450,000$ frames, and over $530,000$ annotated bounding boxes.
As far as we know, this is the largest TIR dataset till now.
Extensive experimental results on the VOT-TIR2015~\cite{VOT-TIR2015}, VOT-TIR2017~\cite{VOTTIR2017}, and PTB-TIR~\cite{PTB-TIR} benchmarks show that the proposed method performs favorably against the state-of-the-art methods.

%contributions
In this work, we make the following contributions:
\begin{itemize}
  \item We propose a feature model comprising TIR-specific discriminative features and fine-grained correlation features for TIR object representation.
        We develop a classification network and a fine-grained aware network to generate the TIR-specific discriminative features and fine-grained correlation features respectively.
        Furthermore, we design a multi-task matching framework for integrating these two features effectively.
  \item We construct a large-scale TIR video dataset with annotations.
        The dataset can be easily used in TIR-based applications and we believe it will contribute to the development of the TIR vision field.
  \item We explore how to better use the grayscale and TIR training datasets for improving a TIR tracking framework and test several strategies.
  \item We conduct extensive experiments on three benchmarks and demonstrate that the proposed algorithm achieves favorable performance against the state-of-the-art methods.
\end{itemize}

\section{Related Work}
\noindent{\textbf{Deep feature based TIR trackers.}} Existing deep TIR trackers usually use the pre-trained feature for representation and combine it with conventional frameworks for tracking.
DSST-tir~\cite{16KTIR} investigates the classification-based deep feature with Correlation Filters (CFs) for TIR tracking and shows that the deep features achieve better performance than the hand-crafted features.
MCFTS~\cite{MCFTS} combines the different layer features of VGGNet~\cite{VGGNet} to construct an ensemble TIR tracker.
LMSCO~\cite{LMSCO} uses the deep appearance and motion features in a structural support vector machine for TIR tracking.
ECO-tir~\cite{ECO-tir} trains a Siamese network on a large amount of synthetic TIR images to extract the deep feature and then combine it with ECO~\cite{ECO} for tracking.
Different from these methods, we propose to learn the TIR-specific discriminative feature and fine-grained correlation feature for representing TIR objects more effectively.

\noindent{\textbf{Matching-based deep trackers.}}
A key issue of the matching-based deep tracker is how to enable its discriminating ability.
Several methods focus on this problem from different aspects.
DSiam~\cite{DSiam} online updates the Siamese network by two linear regression models for adapting to the variation of the object.
CFNet~\cite{CFNet} updates the target template by incorporating a CF module into the network.
SA-Siam~\cite{SA-Siam} learns a twofold matching network by introducing complementary semantic features while FlowTrack~\cite{FlowTrack} combines the optical flow features for matching.
SiamFC-tri~\cite{SiamFC-tri} learns the more discriminative deep features by formulating the triplet relationship using a triple loss.
StructSiam~\cite{StructSiam} learns the fine-grained features for matching using a local structure detector and a context relation model.
RASNet~\cite{RASNet} introduces three kinds of attention mechanisms to adapt the model for online matching.
TADT~\cite{TADT} online selects the target-aware features using two auxiliary tasks for compact matching.
DWSiam~\cite{DeepSiam} uses a deeper and wider backbone network on a Siamese framework to obtain more accurate tracking results.
Different from these methods, we use multiple complementary tasks to learn more powerful TIR features for representing TIR objects.
The proposed multi-task matching network distinguishes TIR objects based on both the inter-class and intra-class differences.

\noindent{\textbf{Multi-task learning.}} When different tasks are sufficient related, multi-task learning can obtain better generalization and benefit all of these tasks.
This is demonstrated in several applications including person re-identification, image retrieval, and object tracking, etc.
MTDnet~\cite{multi-task-reid1} simultaneously takes a binary classification task and a ranking task into account to boost the performance of person re-identification. MSP-CNN~\cite{multi-task-reid2} uses three kinds of task constrains to learn more discriminative features on a Siamese framework for person re-identification.
Cp-mtML~\cite{image-retrieval1} simultaneously learns face identity, age recognition, and expression recognition on heterogeneous datasets for face retrieval.
SiamRPN~\cite{SiamRPN} exploits a classification task and a regression task on a Siamese network to boost the accuracy and efficiency of object tracking.
EDCF~\cite{EDCF} jointly trains a low-level fine-grained matching and high-level semantic matching tasks on a Siamese framework for object tracking.
Different the above methods, we jointly train a classification task, a discriminative matching task, and a fine-grained matching task for robust TIR tracking.

\noindent{\textbf{TIR dataset.}} TIR training dataset is crucial for training a deep TIR tracker.
Most deep TIR trackers only use RGB datasets to train the model, since there is not a proper and large-scale TIR dataset.
This hinders the development of CNNs-based TIR tracking.
To this end, several methods attempt to use TIR data to train a network for tracking.
DSST-tir~\cite{16KTIR} uses a small TIR dataset to train a classification network for feature extraction and then combines it with the DSST tracker for TIR tracking.
ECO-tir~\cite{ECO-tir} explores a Generative Adversarial Network (GAN) to generate synthetic TIR images and then uses them to train a Siamese network for feature extraction. %
The trained model using these synthetic TIR images achieves favorable results.
MLSSNet~\cite{MLSSNet} trains a multi-level similarity based Siamese network on an RGB and TIR dataset simultaneously.
Despite the promising performance they have achieved, the used TIR dataset is not large enough, which hinders them from further improvements.
In this paper, we construct a larger TIR dataset to train the proposed network for adapting the model to the TIR domain.

\begin{figure}[ht]
\begin{center}
\includegraphics[width=0.47\textwidth]{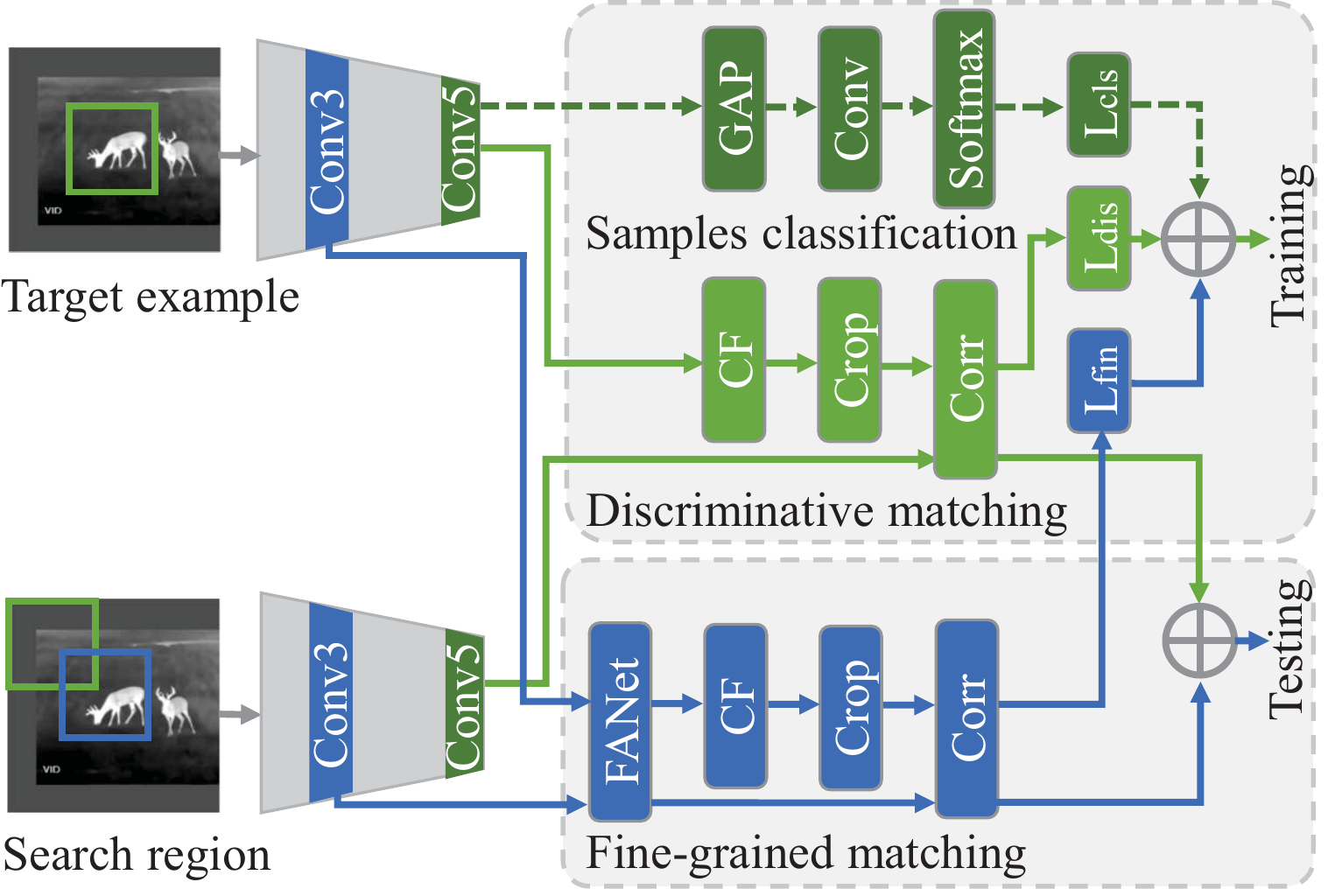}
\end{center}
\caption{Architecture of the proposed Multi-task Matching Network (MMNet). It comprises a shared feature extracted network, a classification branch, a discriminative matching branch, and a fine-grained matching branch.
In this figure, every box denotes a network layer or a subnetwork. Conv, GAP, CF, Corr, and FANet denote the convolution, global average pooling, correlation filter, cross-correlation, and fine-grained aware network (see Fig.~\ref{FANet}), respectively. }
\label{architecture}
\end{figure}

\section{Multi-Task Matching Network}
In this section, we show how to learn TIR-specific features and integrate them in a multi-task matching network for TIR tracking.
First, we present the overall multi-task matching network and introduce the TIR-specific discriminative feature module and the fine-grained correlation feature module.
Then, we introduce the constructed TIR dataset and analyze three multi-domain aggregation learning strategies.
Finally, we give the flow of the tracking algorithm using the proposed model.

\subsection{Multi-task architecture}
We propose a multi-task matching network to integrate the TIR-specific discriminative features and the fine-grained correlation features for TIR tracking.
The network consists of a shared feature extracted network, a discriminative matching branch, a classification branch, and a fine-grained matching branch, as shown in Fig.~\ref{architecture}.
Different from existing trackers using pre-trained features on visual images, the proposed multi-task network uses both TIR-specific discriminative features and fine-grained correlation features for TIR object localization under a matching framework.
In the following, we present the details of each component.

\noindent{\textbf{Discriminative matching.}} Considering tracking efficiency, we use a general matching architecture which is the same as that of CFNet~\cite{CFNet} to perform tracking.
As deeper convolution layers contain more discriminative features, we construct the discriminative matching module on top of the last convolution layer of the shared feature extraction network.
Given a target example $\mathbf{Z}$ and a search image $\mathbf{Y}$ , the discriminative similarity $f_{dis}(\mathbf{Z},\mathbf{Y})$ can be formulated as:
 \begin{equation}\label{semantic}
   f_{dis}(\mathbf{Z},\mathbf{Y})=g(\sigma(\phi_{conv5}(\mathbf{Z})),\phi_{conv5}(\mathbf{Y})),
 \end{equation}
where $\phi_{conv5}(\cdot)$ extracts features using the last convolutional layer of the shared feature extraction network, $g(\cdot,\cdot)$ denotes the cross-correlation operator and $\sigma(\cdot)$ is the CF block which is used to improve the discriminative capacity by online updating the target template.
We adopt a logistic loss to train this branch:
\begin{equation}\label{meanloss}
  \mathcal{L}_{dis}(y,o)=\frac{1}{\mid \mathbf{D} \mid}\sum_{u \in \mathbf{D}}\log(1+\exp(-y[u]o[u])),
\end{equation}
where $\mathbf{D} \in \mathbb{R}^{M\times M}$ is the similarity map generated by Eq.~\ref{semantic}, $o[u]$ denotes the real value of a single target-candidate pair, and $y[u]$ is the ground-truth of this pair.

\noindent{\textbf{TIR-specific discriminative features.}} We use a classification branch as an auxiliary task to obtain the TIR-specific discriminative features and then use them in the discriminative matching branch.
The classification task aiming to distinguish TIR objects belonging to different classes learns the features focusing on the class-level difference.

In the auxiliary network, we first use a global average pooling layer instead of a fully connected layer to avoid the over-fitting problem.
Then, a $1\times 1$ convolution layer is used to adapt the number of the class of the training set.
Finally, we use a cross-entropy loss to train it:
\begin{equation}\label{classloss}
  \mathcal{L}_{cls}(y,p)= -\sum_{k=0}^{K}y_{k}\log p_{k},
\end{equation}
where $y$ is the ground-truth, $p$ is the predicted label, and $K$ denotes a total number of the classes.

\noindent{\textbf{Fine-grained matching.}} The intra-class TIR objects often have a similar visual pattern as they do not have color information.
Coupled with the TIR-specific discriminative branch, we construct a fine-grained matching branch to distinguish intra-class TIR objects.
We note that the fine-grained correlation features are helpful for distinguishing distractors.
We compute the fine-grained correlation feature on a shallow convolution layer since the shallow convolution features mainly contain more detailed information.
The fine-grained similarity can be formulated as:
\begin{equation}\label{structure}
   f_{fin}(\mathbf{Z},\mathbf{Y})=g(\sigma(\omega(\phi_{conv3}(\mathbf{Z}))),\omega(\phi_{conv3}(\mathbf{Y}))),
 \end{equation}
where  $\phi_{conv3}(\cdot)$ extract features using the third convolutional layer of the shared feature extraction network, $\omega(\cdot)$ denotes the proposed fine-grained aware module. We use a logistic loss which is the same with Eq.~\ref{meanloss} to train this branch.

\begin{figure}[t]
\begin{center}
\includegraphics[width=0.48\textwidth]{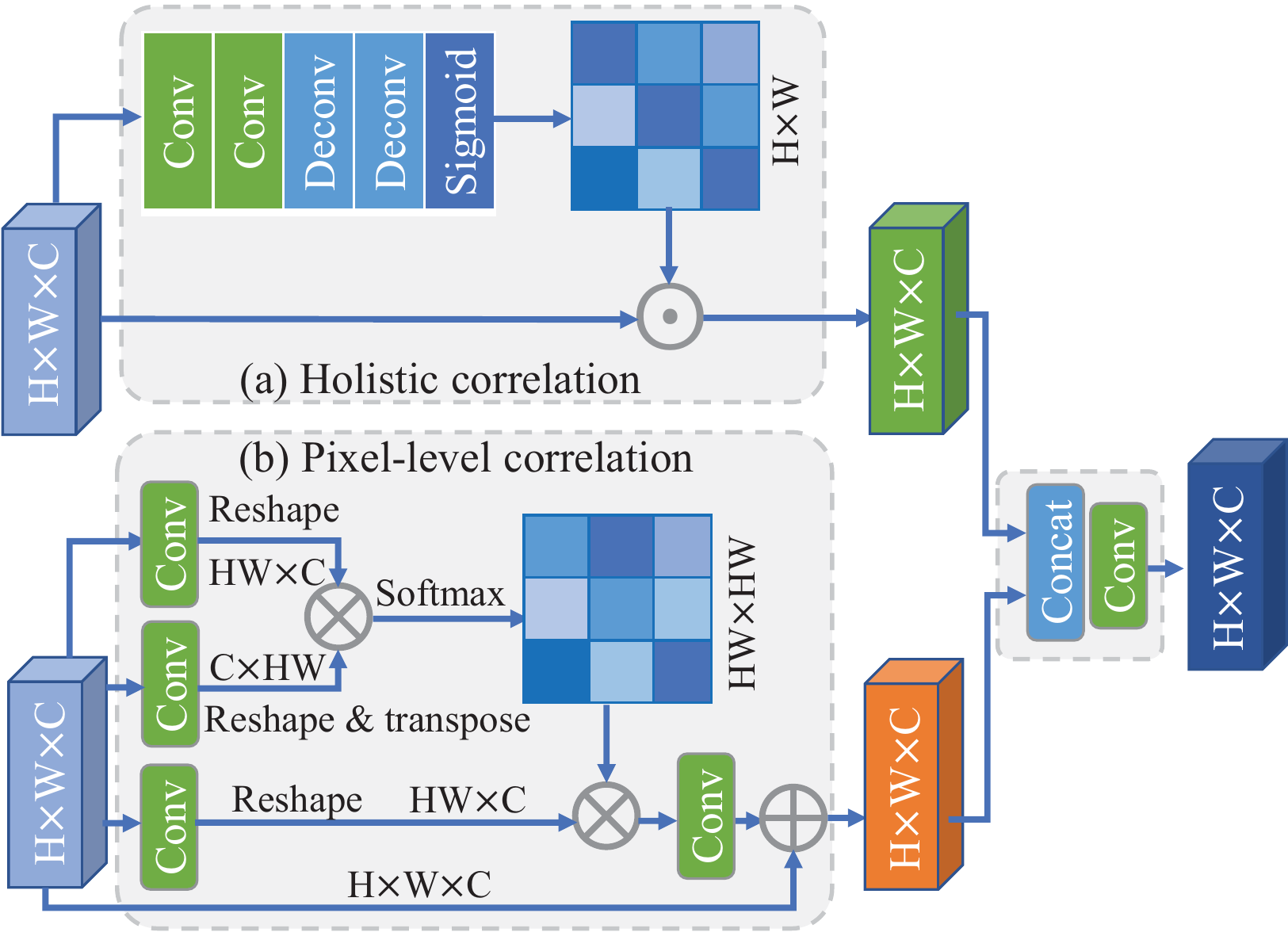}
\end{center}
\caption{Architecture of the proposed Fine-grained Aware Network (FANet). It consists of a holistic correlation module and a pixel-level correlation module. The input and output are a $H\times W\times C$ feature map, $\odot$ denotes the broadcast element-wise multiplication, $\otimes$ denotes the batch matrix multiplication, and $\oplus$ is the broadcast element-wise addition. }
\label{FANet}
\end{figure}

\begin{figure}[t]
\centering
\includegraphics [width=0.48\textwidth]{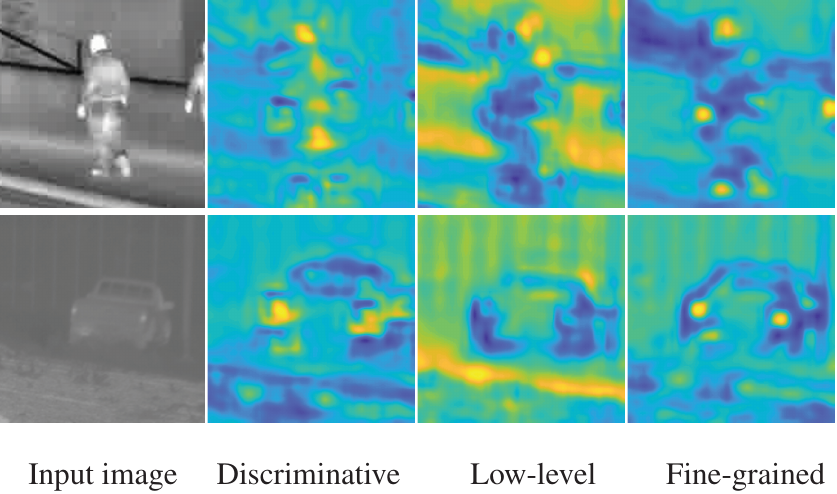}
\caption{Visualization of the TIR-specific discriminative features and fine-grained correlation features. The visualized feature maps are generated by summing all the channels. From left to right, each column shows the original images, the TIR-specific discriminative features (Conv5), the low level features (Conv3), and the learned fine-grained correlation features from Conv3 respectively. This figure shows that TIR-specific discriminative feature is too coarse to achieve accurate localization , while the fine-grained correlation feature map focuses on local prominent regions which contributes to accurate localization. }
\label{featuremap}
\end{figure}

%fine-grained aware network
\noindent{\textbf{Fine-grained correlation features.}}
To get the fine-grained correlation features, we design a fine-grained aware network which consists of a holistic correlation module and a pixel-level correlation module.
Fig.~\ref{FANet} depicts the architecture.
Given an input feature map $ \mathbf{X}\in \mathbb{R}^{H\times W\times C}$, the fine-grained aware module can be formulated as:
\begin{equation}\label{SA}
  \omega(\mathbf{X}) = f_{c}(\varphi_{h}(\mathbf{X}),\varphi_{p}(\mathbf{X})),
\end{equation}
where $\varphi_{h}(\cdot)$ denotes the holistic correlation module which formulates the relationship between local regions,
$\varphi_{p}(\cdot)$ denotes the pixel-level correlation module which is used to formulate the relationship between all feature units, and $f_{c}(\cdot,\cdot)$ is cascaded by a concat and a $1 \times 1$ convolutional layers, which integrates these two complementary correlations.
Fig.~\ref{featuremap} compares the TIR-specific discriminative feature and the fine-grained correlation feature using visualizations of the feature maps.

To formulate the relationships between local regions, we use an encoder-decoder architecture based on a self-attention mechanism.
We first exploit two large convolution kernels to find out discriminative local regions.
Then, we use two deconvolution layers to locate them.
After that, a correlation map is generated using a Sigmoid activation function.
The map denotes the importance of every local region.
Finally, we weight the original feature map using this correlation map for making it focus on the local discriminative regions.
The weighed feature map is computed as:
\begin{equation}\label{Satt}
\varphi_{h}(\mathbf{X}) = \mathbf{X} \odot \frac{\exp(\mathbf{W}\mathbf{X})}{\exp(\mathbf{W}\mathbf{X})+1},
\end{equation}
where $\mathbf{W}$ denotes the transform matrix which is constituted by two convolution and two deconvolution layers.

As pixel-level context information is crucial for representing TIR objects, we exploit a pixel-level correlation module to formulate the relationships between every feature unit for obtaining more fine-grained correlation information.
The pixel-level correlation model is similar to the non-local network~\cite{NLNet} which captures long-range dependencies.
Specifically, we first formulate the pixel-level relationships with a spatial correlation map $\mathbf{S}\in \mathbb{R}^{HW\times HW}$, which is computed as:
\begin{equation}\label{nl attention map}
  s_{ij} = \frac{\exp(\mathbf{W}_{q}\mathbf{x}_{i}\bigotimes \mathbf{W}_{k}\mathbf{x}_{j})}{\sum_{n=1}^{N}\exp(\mathbf{W}_{q}\mathbf{x}_{i}\bigotimes \mathbf{W}_{k}\mathbf{x}_{n})},
\end{equation}
where $s_{ij} \in \mathbf{S}$ denotes the relationship between the $i$-th feature unit and the $j$-th feature unit, $\mathbf{W}_{q}$ and $\mathbf{W}_{k}$ represent the two $1 \times 1$ convolutional layers respectively, $\mathbf{x}_{i}$ is the $i$-th feature unit in $\mathbf{X}$, and $\mathbf{X}=\{\mathbf{x}_{i}\}_{i=1}^{N}$, where $N=HW$.
Then, we apply this correlation map on the input feature map to obtain the pixel-level correlation feature which can be formulated as:
\begin{equation}\label{Snl}
  \mathbf{S}_{p}= \sum_{j=1}^{N}\sum_{i=1}^{N}s_{ij}(\mathbf{W}_{g}\mathbf{x}_{j}),
\end{equation}
where $\mathbf{W}_{g}$ is a transform matrix which is implemented with a $1\times 1$ convolutional operator. Finally, we perform a weighted sum to the pixel-level correlation feature map and the origin low-level feature map to get the comprehensive correlation feature map using a residual-like connection:
\begin{equation}\label{Sb}
\varphi_{p}(\mathbf{X})= \mathbf{X} + \delta \mathbf{S}_{p},
\end{equation}
where $\delta$ is a scale factor which can be learned automatically.

\subsection{TIR dataset}
To better adapt the proposed model to the TIR domain, we construct a large-scale TIR dataset for training the proposed network.
The dataset consists of 30 classes and over $1100$ sequences.
We annotate the object in every frame of each sequence with bounding box and class labels using a semi-automatic tracking application according to the VID2015~\cite{ILSVRC} style.
Some examples of the annotated videos and comparison with existing tracking datasets are shown in the \textbf{supplementary material}.
The dataset includes more than $450,000$ frames and $530,000$ bounding boxes.
Since most of our sequences are collected from Youtube website, it has a wide range of shotting devices, shotting scenes, and shotting view angles which ensure the diversity of the dataset.
For examples, there are four kinds of shotting devices and view angles: hand-held, vehicle-mounted, surveillance (static), and drone-mounted.
We store these TIR images with a white-hot style and an 8 bits depth.

\subsection{Multi-domain aggregation}

%training
We find that the grayscale image sample can provide rich detailed information, e.g., texture and structure, which is helpful to the TIR tracking task.
As such, we explore to use both the grayscale and TIR domains to boost the TIR tracking performance.
To find an effective way to combine them, we test three multi-domain aggregation learning strategies.
\begin{itemize}
  \item \textbf{Re-training}. We first train the proposed network on the VID2015~\cite{ILSVRC} grayscale dataset with a multi-task loss:
\begin{equation}\label{mutti-task loss}
 \mathcal{L} = \lambda_{1}\mathcal{L}_{dis} + \lambda_{2}\mathcal{L}_{cls} + \lambda_{3}\mathcal{L}_{fin},
\end{equation}
where $\mathcal{L}_{fin}$ denotes the fine-grained similarity loss which is same as $\mathcal{L}_{dis}$.
Then, we re-train the overall network on the TIR dataset.
  \item \textbf{Fine-tuning.} We also use the trained model on VID2015 as initial parameters of the proposed network and freeze the first three layers of the shared feature extracted network and the fine-grained matching branch for retaining the detail information.
      Then, we use a smaller learning rate to fine-tune the network on the TIR dataset.
  \item \textbf{Mix-training.} We first mix the VID2015 and TIR dataset together and get a new mixed dataset.
      Then, we freeze the classification branch and train the proposed network from scratch on the mixed dataset.
\end{itemize}
In the Ablation studies section, we report and analyze the results of each strategy.

\subsection{Tracking process}
Once the multi-task matching network is learned, we prune the classification branch and use the rest part for online TIR tracking without updating.
Fig.~\ref{architecture} shows the testing framework.
Given a target instance $\mathbf{Z}_{t-1}$ at the $(t-1)$-th frame and a search image $\mathbf{Y}_{t}$ at the $t$-th frame, the prediction in the $t$-th frame can be computed as:
\begin{equation}\label{trackingprocess}
   \hat{\mathbf{y}}_{t,i}= \mathop{\arg\max}_{\mathbf{y}_{t,i}} f_{dis}(\mathbf{Z}_{t-1},\mathbf{Y}_{t})+f_{fin}(\mathbf{Z}_{t-1},\mathbf{Y}_{t}),
\end{equation}
where $\mathbf{y}_{t,i} \in \mathbf{Y}_{t}$ is the $i$-th candidate in the search region $\mathbf{Y}_{t}$. We use a scale-pyramid mechanism~\cite{Siamese-fc} to estimate the size change of the object.

\section{Experimental Results}
\subsection{Implementation details}
We conduct the experiment using the MatConvNet~\cite{matconvnet} toolbox on a PC with an i7 4.0 GHz CPU and a GTX-1080 GPU.
The average speed is about 19 FPS.
We remove all the paddings of AlexNet~\cite{AlexNet} and use it as the base feature extractor.
We train the proposed network using a Stochastic Gradient Descent (SGD) method with the batch size of $8$ and momentum of $0.9$.
At the first stage, we train the network with 60 epochs on the VID2015 dataset and the learning rate exponentially decays from $10^{-2}$ to $10^{-5}$.
We set $\lambda_{1}=\lambda_{2}=\lambda_{3}=1$ of Eq.~\ref{mutti-task loss} at all training stages.
At the re-training and fine-tuning stages, we train the network 30 epochs with the learning rate exponentially decays from $10^{-3}$ to $10^{-5}$ on the constructed TIR dataset.
In the mix-training process, we train the network 70 epochs using the same parameters with the training on VID2015 dataset.

\begin{table*}[t]
\centering
\caption{Ablation studies of the proposed model on the VOT-TIR2015 and VOT-TIR2017 benchmarks. Dis, Cls, Fine-Hc, and Fine-Pc denote the discriminative matching branch, the classification branch, the fine-grained matching with the holistic correlation module, and the fine-grained matching with the pixel-level correlation module respectively.}  %The up arrow and down arrow denote the bigger or smaller value is, the better corresponding performance has. }
\begin{tabular}{cccc|rrr|rrr}
\hline
 \multicolumn{4}{c|}{Tracker} & \multicolumn{3}{c|}{VOT-TIR2015} & \multicolumn{3}{c}{VOT-TIR2017} \\
\hline
 Dis & Cls & Fine-Hc & Fine-Pc  & EAO $\uparrow$ & Acc $\uparrow$ & Rob $\downarrow$  & EAO $\uparrow$  & Acc $\uparrow$ & Rob $\downarrow$ \\
\hline
 \checkmark & & & & 0.282 & 0.55 & 2.82 & 0.254 & 0.52 & 3.45 \\
\checkmark & \checkmark & &  & 0.307 & 0.51 & 2.41 & 0.274 & 0.52 & 3.20 \\
\checkmark & \checkmark & \checkmark &  & 0.322 & 0.58 & \textbf{2.14} & 0.296 & 0.55 & 2.96 \\
\checkmark & \checkmark & &\checkmark  & 0.326 & 0.57 & 2.30 & 0.279 & 0.56 & 3.24 \\
\checkmark & \checkmark & \checkmark & \checkmark    & \textbf{0.332} &\textbf{0.60}  &2.26  &\textbf{0.320} & \textbf{0.58} & \textbf{2.91} \\
\hline
\end{tabular}
\label{ablation}
\end{table*}

\subsection{Ablation studies}
\label{ablationstudies}

\noindent{\textbf{Datasets.}} The VOT-TIR2015~\cite{VOT-TIR2015} and VOT-TIR2017~\cite{VOTTIR2017} benchmarks are widely used for evaluating TIR trackers. These two datasets contain six kinds of challenges, such as dynamics change, camera motion, and occlusion.
Each challenge has a corresponding subset which can be used to evaluate the ability of a tracker to handle the challenge.
In addition to the VOT-TIR2015 and VOT-TIR2017 datasets, we also use a TIR pedestrian tracking dataset, PTB-TIR~\cite{PTB-TIR}, to evaluate the proposed algorithm.
PTB-TIR is a recently published tracking benchmark that contains 60 sequences with 9 different challenges, such as background clutter, occlusion, out-of-view, and scale variation.

\noindent{\textbf{Evaluation criteria.}} VOT-TIR2015 and VOT-TIR2017 use Accuracy (Acc) and Robustness (Rob)~\cite{ARplot} to evaluate the performance of a tracker from two aspects.
Accuracy is the average overlap rate between the predicted bounding box and the ground truth bounding box.
Robustness denotes the average frequency of tracking failure on the overall dataset. %
In addition, Expected Average Overlap (EAO) is often used to evaluate the overall performance of a tracker, which is computed based on Acc and Rob.
PTB-TIR uses the Precision (Pre) and Success (Suc) plots to evaluate the performance of a tracker.
The precision plot measures the percentage of frames whose Center Location Error (CLE) is within a given threshold (20 pixels), the success plot measures the percentage of frames whose Overlap Ration (OR) is larger than a given threshold.
The Area Under the Curve (AUC) of the precision and success plots are often used to rank methods.

\noindent{\textbf{Network architecture.}}
Table~\ref{ablation} shows the results of ablation study.
From the first two rows, we can see that the classification branch (Cls) improves the robustness of the tracker with more than $2\%$ gains of EAO score on both benchmarks.
This shows the effectiveness of the TIR-specific discriminative features.
From the second to fourth rows, we can see that the fine-grained matching branch using the holistic correlation module (Fine-Hc) improves the accuracy by $7\%$ and $3\%$ on these two benchmarks respectively, while the fine-grained matching branch using the pixel-level correlation module (Fine-Pc) improves the accuracy by $6\%$ and $4\%$ on these two benchmarks respectively.
The last row shows that the fine-grained matching branch using both the holistic and pixel-level correlation modules further improves the accuracy by more than $2\%$ on both benchmarks.
We attribute these gains to the fine-grained correlation features, which are effective in distinguishing similar objects, and the complement advantages of the holistic correlation and pixel-level correlation modules, which provide more powerful features for target localization.

\begin{table}[t]
\centering
\caption{Comparison of the different models using two single-domain learning methods and three multi-domain aggregation learning strategies on the VOT-TIR2015 and PTB-TIR benchmarks. }
\begin{tabular}{l|rrr|rr}
\hline
   & \multicolumn{3}{c|}{VOT-TIR2015}  &\multicolumn{2}{c}{PTB-TIR} \\
\hline
 Strategy & EAO $\uparrow$ & Acc $\uparrow$ & Rob $\downarrow$  & Pre $\uparrow$ & Suc $\uparrow$   \\
\hline
 Only-VID &0.332 &0.60 &2.26 & 0.661 & 0.502   \\
 Only-TIR &0.311 &0.55 &2.47 &0.694  &0.519   \\
\hline
 Re-training &0.300 &0.58 &2.37 &0.730  &0.521    \\
  Fine-tuning &0.322 &0.58 &2.16 &0.729   &0.525     \\
  Mix-training &\textbf{0.344} &\textbf{0.61} & \textbf{2.09} & \textbf{0.759}  &\textbf{0.539}     \\
\hline
\end{tabular}
\label{MDS}
\end{table}

\noindent{\textbf{Multi-domain aggregation.}} Table~\ref{MDS} shows the results of the proposed model using different training strategies.
Compared with only training on the VID2015 dataset (Only-VID), the mix-training learning strategy achieves a $1.2\%$ EAO score gain on VOT-TIR2015 and a $3.7\%$ success rate gain on PTB-TIR.
Compared with only training on the TIR dataset (Only-TIR), the mix-training strategy also improves the EAO score by $3\%$ on VOT-TIR2015 and the success rate by $2\% $ on PTB-TIR.
These results demonstrate that the mix-training can make full use of the property of grayscale and TIR images to get more powerful features for TIR tracking.
Compared with Only-VID, the fine-tuning strategy achieves a $2.3\%$ gain of the success rate and a $6.8\% $ gain of precision on PTB-TIR.
It also improves the robustness on VOT-TIR2016.
These results demonstrate that the fine-grained feature learned from the grayscale dataset are useful to TIR tracking.
The re-train with the TIR dataset does not improve the performance significantly on both datasets.
This is because TIR images lack detailed features for precise locating.

\begin{figure}[t]
\centering
\includegraphics [width=0.47\textwidth]{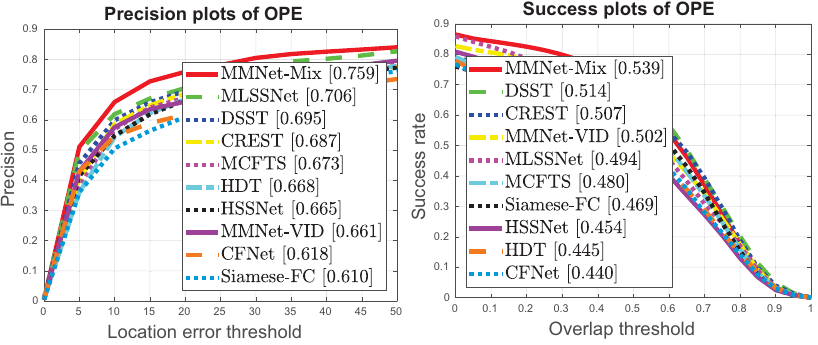}
\caption{Comparison of ten trackers on the PTB-TIR benchmark. }
\label{PTB-TIR}
\end{figure}

\begin{figure}[t]
\centering
\includegraphics [width=0.47\textwidth]{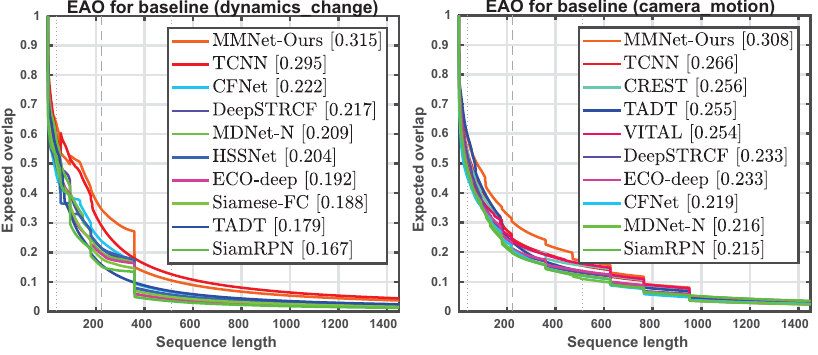}
\caption{EAO scores of the top ten trackers on two challenges of the VOT-TIR2017 benchmark. }
\label{VOT-TIR2017-Att}
\end{figure}

\begin{table*}[t]
\centering
\caption{Comparison of our tracker and the state-of-the-art methods on VOT-TIR2017 and VOT-TIR2015. The bold and underline denote the best and the second-best scores, respectively. The notation ``*" denotes the speed is reported by the authors.}
\begin{tabular}{l|l|rrr|rrr|r}
\hline
 \multirow{2}{2.5cm}{Category} & \multirow{2}{2.5cm}{Tracker} & \multicolumn{3}{c|}{VOT-TIR2017} & \multicolumn{3}{c|}{VOT-TIR2015} & Speed\\ \cline{3-9}
 & & EAO $\uparrow$ & Acc $\uparrow$ & Rob $\downarrow$  & EAO $\uparrow$ & Acc $\uparrow$ & Rob $\downarrow$ & FPS\\
\hline
 \multirow{2}{2.5cm}{Hand-crafted feature based CF} &SRDCF~\cite{SRDCF} & 0.197 &0.59 &3.84 &0.225 &0.62 &3.06 & 12.3 \\
  &Staple-TIR~\cite{VOT-TIR2016}& 0.264 & \textbf{0.65} & 3.31   &- &- &-  &80.0* \\
\hline
  \multirow{5}{2.5cm}{Deep feature based CF}&MCFTS~\cite{MCFTS}  & 0.193 & 0.55 & 4.72  &0.218 &0.59 & 4.12  & 4.7\\
  &HDT~\cite{HDT}& 0.196 & 0.51 & 4.93  &0.188 &0.53 &5.22   & 10.6\\
  &deepMKCF~\cite{DeepMKCF}  & 0.213 & 0.61 & 3.90 &- &- &-  &5.0*\\
  &CREST~\cite{CREST}  & 0.252 & 0.59 & 3.26 &0.258 &0.62 &3.11  &0.6\\
  &DeepSTRCF~\cite{DeepSTRCF}             &0.262  &0.62  &3.32 &0.257 &0.63 &2.93 &5.5\\
  &ECO-deep~\cite{ECO}          &0.267  &0.61 &\underline{2.73}  &0.286 &\underline{0.64} &2.36 &16.3\\
  \hline
    \multirow{3}{2.5cm}{Other deep tracker}&MDNet-N~\cite{VOT-TIR2016}  & 0.243 & 0.57 & 3.33   &-  &- &-  & 1.0*\\
   &VITAL~\cite{VITAL}          &0.272  &\underline{0.64} &\textbf{2.68} &0.289 &0.63 &\underline{2.18} &4.7\\
   &TCNN~\cite{TCNN}       &\underline{0.287} & 0.62 &2.79 &- &- &- &1.5*\\
\hline
  \multirow{6}{2.5cm}{Matching based deep tracker}&Siamese-FC~\cite{Siamese-fc}  & 0.225 & 0.57 & 4.29 &0.219 &0.60 &4.10 &66.9   \\
  &SiamRPN~\cite{SiamRPN}         & 0.242 &0.60  &3.19  &0.267 &0.63 &2.53 & 160.0* \\
  &CFNet~\cite{CFNet}   & 0.254 & 0.52 & 3.45 &0.282 &0.55 &2.82  & 37.0\\
  &HSSNet~\cite{HSSNet}   & 0.262 & 0.58 &3.33 &0.311 &\textbf{0.67} &2.53 &10.0*\\
  &TADT~\cite{TADT} &0.262 &0.60 &3.18 &0.234 &0.61 &3.33 & 42.7 \\
  &MLSSNet~\cite{MLSSNet}    & 0.278 & 0.56  & 2.95 &\underline{0.316} &0.57 &2.32 &18.0\\
  &MMNet (Ours) & \textbf{0.320}& 0.58 & 2.91 &\textbf{0.344} &0.61 &\textbf{2.09} & 18.9 \\
\hline
\end{tabular}
\label{SOTA}
\end{table*}

\subsection{Comparison with state-of-the-arts}
\noindent{\textbf{Compared trackers.}} We compare the proposed method with the state-of-the-art trackers including hand-crafted feature based correlation filter trackers, such as DSST~\cite{DSST}, SRDCF~\cite{SRDCF}, and Staple-TIR~\cite{VOT-TIR2016}; the deep feature based correlation filter trackers, such as HDT~\cite{HDT}, deepMKCF~\cite{DeepMKCF}, CREST~\cite{CREST},  MCFTS~\cite{MCFTS}, ECO-deep~\cite{ECO}, and DeepSTRCF~\cite{DeepSTRCF}; matching based deep trackers, such as CFNet~\cite{CFNet}, Siamese-FC~\cite{Siamese-fc}, SiamRPN~\cite{SiamRPN}, HSSNet~\cite{HSSNet}, MLSSNet~\cite{MLSSNet}, and TADT~\cite{TADT}; and other deep trackers, such as TCNN~\cite{TCNN}, MDNet-N~\cite{VOT-TIR2016} and VITAL~\cite{VITAL}.

\noindent{\textbf{Results on PTB-TIR.}} Fig.~\ref{PTB-TIR} shows that the proposed algorithm achieves the best success rate of $0.539$ and precision of $0.759$ on PTB-TIR.
Compared with CFNet which just uses a single matching branch, the proposed method (MMNet-Mix) achieves a $10\%$ relative gain of the success rate.
Although the proposed method (MMNet-VID) is not trained on the TIR dataset, it also improves the success rate by $6\%$.
This demonstrates the effectiveness of the proposed network architecture and the mix-training learning strategy.
Compared with the correlation filter based deep trackers, the proposed method obtains a better success rate.
We attribute the good performance to the specifically designed TIR feature model and the constructed large-scale TIR dataset.

\noindent{\textbf{Results on VOT-TIRs.}} As shown in Table~\ref{SOTA}, the proposed method (MMNet) achieves the best EAO scores of $0.320$ and $0.344$ on VOT-TIR2017 and VOT-TIR2015, respectively.
Compared with other matching based deep trackers, the proposed multi-task matching network learns more effective TIR features for matching.
Although TADT online selects more compact and target-aware features from a pre-trained CNN for matching, the proposed method still obtains a better performance on both benchmarks.
Compared with the best correlation filter based deep tracker, ECO-deep, which uses the classification-based pre-trained feature, the proposed method obtains better robustness on VOT-TIR2015.
This benefits from the learned fine-grained correlation features which help the multi-task matching network distinguish similar distractors.
Compared with the best deep tracker, TCNN, which uses multiple CNNs to represent objects, the proposed method achieves a better performance on VOT-TIR2017 while running faster.
We attribute the good performance to the proposed TIR-special feature model which is more effective in representing TIR objects.
Fig.~\ref{VOT-TIR2017-Att} shows that our method achieves the best EAO on the dynamic change and camera motion challenges of VOT-TIR2017.
Compared with the second best matching based tracker, CFNet, the proposed method achieves a $9.3\%$ EAO score gain on the dynamics change challenge.
This shows that the proposed TIR-special feature model is more robust to the appearance variation of the target.
Furthermore, the proposed method achieves a higher EAO score than the second best method (TCNN) by $4.2\%$ on the camera motion challenge.
Some more attribute-based results can be found in the \textbf{supplementary material}.
These results demonstrate the effectiveness of the proposed algorithm.

\section{Conclusions}
In this paper, we propose to learn a TIR-specific feature model for robust TIR tracking.
The feature model includes a TIR-specific discriminative feature module and a fine-grained correlation feature module.
To use these two feature models simultaneously, we integrate them into a multi-task matching framework.
The TIR-specific discriminative features, generated with an auxiliary multi-classification task, are able to distinguish inter-class TIR objects.
The fine-grained correlation features are obtained with a fine-grained aware network consisting of a holistic correlation module and a pixel-level correlation module.
These two kinds of features complement each other and distinguish TIR objects in the levels of inter-class and intra-class, respectively.
In addition, we develop a large-scale TIR training dataset for adapting the model to the TIR domain, which can be also easily applied to other TIR tasks.
Extensive experimental results on three benchmarks demonstrate that the proposed method performs favorably against the state-of-the-art methods.

\section{Acknowledgment}
This work is supported by the National Natural Science Foundation of China (Grant No. 61672183), by the Natural Science Foundation of Guangdong Province (Grant No.2015A030313544), by the Shenzhen Research Council (Grant No. JCYJ20170413104556946, JCYJ20170815113552036), and by the project "The Verification Platform of Multi-tier Coverage Communication Network for oceans (PCL2018KP002)".

\bibliographystyle{aaai}

\end{document}